# Weight mechanism: adding a constant in concatenation of series connect

Xiaojie Qi[1]

*Abstract*—It's a consensus that feature maps in the shallow layer are more related to image attributes such as texture and shape, whereas abstract semantic representation exists in the deep layer. Meanwhile, some image information will be lost in the process of the convolution operation. Naturally, the direct method is combining them together to gain lost detailed information through concatenation or adding. In fact, the image representation flowed in feature fusion can't match with the semantic representation completely, and the semantic deviation in different layers also destroy the information purification, that leads to useless information being mixed into the fusion layers. Therefore, it's crucial to narrow the gap among the fused layers and reduce the impact of noises during fusion. In this paper, we propose a method named weight mechanism to reduce the gap between feature maps in concatenation of series connection, and we get a better result of 0.80% mean Intersection over Union (mIoU) improvement on Massachusetts building dataset by changing the weight of the concatenation of series connection in residual U-Net. Specifically, we design a new architecture named fused U-Net to test weight mechanism, and it also gains 0.12% mIoU improvement.

## I. INTRODUCTION

SEMANTIC segmentation (SS) means classifying each pixel of the image, and the outcome of SS can reflect the boundary of objects in the picture, which is extremely important in precise automatic driving and satellite images analyzing, etc.

As a promising method to extract image information effectively and accurately, SS is also one of the most challenging tasks in computer vision. The basic contradiction in SS models is that when the depth of the convolutional neural network (CNN) increases, the feature maps in the deep layer own more semantic representation and fewer image representation such as shape and texture. It signifies high-resolution feature maps exist in the shallow layer, and high-level representation exists in the deep layer. In the meantime, a part of image information will be filtered in the process of nonlinear transformation. Obviously, the outcome will be rough if the SS model just relies on the semantic representation to accomplish the decoding process.

In order to deal with this issue, a straightforward method is to combine high-resolution and high-level feature maps together. Directly, it not only provides abundant information that contains different levels of representation for the next convolution operation, but also enhances the expression capability of CNN by increasing channels for concatenation or enrich the feature maps for adding. However, the information extracted by the shallow layer can't match with the feature maps filtered by the deep layer exactly. Afterwards, the redundant information mixed in feature fusion confuses the processing of filter and decreases the accuracy of extracted information of SS models. Another idea is that adding a gate in the fusion path to filter out the useless feature maps and enlarge the beneficial information. Usually the elements of the gate contain a convolution layer and an activation function. The parameters of the gate are dynamic in the training process, and the gate is nonlinear. Many works [1] - [3] have proved that it's a useful skill to optimize models. Even so, the gate still has numerous parameters and complex propagation. Hence, a simple fusion strategy is required to collect information efficiently in feature fusion.

In this paper, we propose a method named weight mechanism, and it improves the accuracy of the Massachusetts building dataset [4] in residual U-Net, even better than gating mechanism. The main idea is that the weight can reduce the value of a part of feature maps in concatenation, and then the multiplied convolutional weight will enlarge the useful feature maps automatically. The key point is to keep the weight in weight mechanism relatively independent of the weight in convolutional layers. Otherwise, it will just cause terrible initialization of the layers and then decrease the accuracy of the model outcome. The experiments show that the weight $\alpha$ illustrated in Fig. 1(a) works well, but the weight $\beta$ illustrated in Fig. 1(b) will mess the model result. What's more, we design a more complex architecture named fused U-Net. In comparison experiments with limited training times, it still achieved a 0.12% mIoU improvement. It shows that weight mechanism has broad application space.

## II. RELATE WORK

In this section, we review feature fusion methods for SS from two categories, i.e., multilevel and gated feature fusion.

### A. Multilevel feature fusion:

Multilevel feature fusion means different level of feature maps are combined by adding or concatenation. In residual block [5], the way to keep the feature information is that adding the input and the output of the block together. Given that the residual block can be trained easily, it's widely used in CNNs. Moreover, He *et al.* [6] proved that if multiplying parameters in the adding path like Fig. 1(c), it will decrease the accuracy of classification in CIFAR-100 dataset, as same as gates [1] and dropout [7]. Similarly, it is comprehensively utilized to concatenate two different level of feature maps in classic SS models, such as FCN [8] and U-Net [9]. The architecture of U-

[1] Xiaojie Qi's e-mail: qixiaojieqingdao@foxmail.com.

Net is simple but effective, it can extract quite accurate results in the case of limited datasets. Therefore, we choose U-Net combined with residual block as the competitive baseline model.

With the dataset being more complex, the capacity of combining two different level of feature maps is not enough to keep the exhaustive information of images. Consequently, it is natural to consider more complicated methods to do feature fusion. Huang *et al.* [10] proposed dense connectivity which adds direct connections from each layer to all subsequent layers. UNet++ [11] and DenseASPP [12] follow the idea of dense connectivity and combine it with other tricks such as U-Net architecture and atrous spatial pyramid pooling (ASPP) [13]. Zhao *et al.* [14] proposed pyramid pooling module, and it enhances global contextual capacity through combining different level of pyramid feature map together. Contrast to link different level of feature maps as much as possible, Zhou *et al.* [15] just concatenated the neighbor representation to exchange information, enabling them to integrate local and contextual information efficiently. Sun *et al.* [16] proposed HRNet. It can keep high-resolution representation, and fuses every level of pyramid feature map with each other as well, and it works well in variable computer vision tasks such as human pose estimation and SS.

*B. Gated feature fusion:*

Comparing to fuse multilevel feature maps completely, adding gate operation will pick up useful feature maps for specific feature fusion. Inspired by LSTM [17], highway networks [18] applies gating mechanism in SS model, it can strengthen the valuable information passing through the deep convolution layers. Xu *et al.* [2] utilized the gating mechanism to guide the message passing for different tasks synchronously. Li *et al.* [3] proposed a GFF model to fuse useful information simultaneously.

Our method is inspired by the above ideas, and multiplying the weight $\alpha$ with $X_1$ in Fig. 1(a) can be seen as a simple linear transform process. The weight is just a constant, and it can scale the representation effectively in concatenation.

## III. METHOD

In this section, we first review the classic feature fusion and present it with mathematics. Then, we introduce the weight mechanism for concatenation of series connection and prove how it works for feature fusion.

*A. Feature Fusion*

For feature fusion, concatenating or adding the feature maps together is the naive method, the whole process can be described as follows:

$$\widetilde{X}_l = \begin{cases} concat(X_1, ..., X_L) \\ \sum_{i=1}^{L} X_i \end{cases} \quad (1)$$

Where $\widetilde{X}_l \in R^{H_l \times W_l \times C_l}$ is the fused feature map for $l$th level. $X_i \in R^{H_i \times W_i \times C_i}$ denotes the $i$th feature map before fusion, $i \in \{1, ..., L\}$. Specially, the process can be seen as simple feature fusion when $L=2$.

As to gated feature fusion, it's still a developing trick that how to design and utilize gating mechanism. Nevertheless, there exists a commonality that the gate is dynamic and nonlinear. A simple application in GFF is defined as follows:

$$G_i = sigmoid(w_i * X_i) \quad (2)$$

$$\widetilde{X}_l = (1+G_l) \odot X_l + (1-G_l) \odot \sum_{i=1, i \neq l}^{L} G_i \odot X_i \quad (3)$$

Where $w_i \in R^{1 \times 1 \times C_i}$ denotes the weights for $i$th level feature map $X_i$. $G_i$ denotes the $i$th Gate for the feature map $X_i$. $i \in \{1,...,L\}$ and $i \neq l$. $\odot$ denotes element-wise multiplication.

*B. Weight Mechanism for Concatenation*

Unlike the naive method and dynamic gating mechanism, we propose a novel method named weight mechanism to scale the information mixing procedure.

$$\widetilde{X}_l = concat(\alpha * X_1, X_L) \quad (4)$$

What's crucial to make weight mechanism in concatenation work is the position of the weight multiplies with the feature map, such as the situation illustrated in Fig. 1(a).

In Fig. 1(a), the whole forward propagation can be clarified as follows:

$$X_1 = H(X_0) = W_1 \otimes X_0 \quad (5)$$
$$X_2 = H(X_1) = W_2 \otimes X_1 \quad (6)$$
$$\widetilde{X}_l = concat(\alpha * X_1, X_2)$$
$$= concat(\alpha * W_1 \otimes X_0, W_2 \otimes (W_1 \otimes X_0)) \quad (7)$$

Where $H(\cdot)$ denotes convolution process, active function and normalization are ignored for simplification. $\otimes$ denotes the convolution operation. $W_i \in R^{H_i \times W_i \times C_i}$ is the convolutional weight for $X_i$. $\alpha$ is the weight to control the scale of mixed feature maps.

In (7), we can see that $\alpha$ can't change $W_1$ directly but affect $W_1$ indirectly by backward propagation of $\widetilde{X}_l$. Hence, choosing a reasonable $\alpha$ can scale $X_1$ to match with $X_2$.

However, in Fig. 1(b), the whole forward propagation is computed as:

$$X_1 = H(X_0) = W_1 \otimes X_0 \quad (8)$$
$$X_2 = H(X_0) = W_2 \otimes X_0 \quad (9)$$
$$\widetilde{X}_l = concat(\beta * X_1, X_2)$$
$$= concat(\beta * W_1 \otimes X_0, W_2 \otimes X_0)$$
$$= concat(W_1^{'} \otimes X_0, W_2 \otimes X_0) \quad (10)$$

In (10), $\beta * W_1$ can be seen as $W_1^{'}$, which means $W_1$ absorbs $\beta$. Namely, $\beta$ just changes the initialization of $W_1$, and $\beta$ will affect the whole propagation in the training process directly. Similarly, $\beta$ in Fig. 1(d) can't make feature fusion better but only ruins the weight initialization stably.

It's noteworthy that the batch normalization (BN) [18] and ReLU activation function are ignored in the forward propagation, but it's obvious that the weight $\beta$ influences the weight and bias of BN more seriously than $\alpha$ like $W_1$, which can be observed by the experiments in section IV.B.

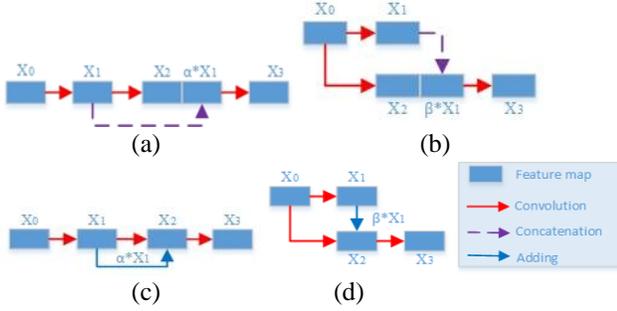

Fig. 1 Illustration of different types of feature fusion. (a) and (c) show the weighted concatenation and adding of series connection. They contain two conditions: $X_2$ can be calculated by $X_1$ through convolution operation; the weight multiples with $X_1$. (b) and (d) show the weighted concatenation and adding of parallel connection. They contain two conditions: $X_2$ can't be calculated by $X_1$ through convolution operation, the weight multiples with $X_1$ or $X_2$. Notice that if multiplying $\alpha$ with $X_2$ rather than $X_1$ in (a) or (c), it becomes a parallel connection, and the feature maps at the same level with $X_2$ are zeros.

### C. Network Architecture

Our baseline model is based on U-Net, and in the down-sample and up-sample processes, the residual block is chosen as backbone for feature extraction. The structure of residual block is same as bottleneck block [5]. In the down-sample process, we replace max pooling with convolution layers that the stride is 2. Moreover, the method of up-sample is bilinear interpolation. The detailed architecture parameters of baseline model are illustrated in Fig. 2

Considering the fairness, in the comparative experiment, what we changed is just the weight $\alpha_i$ or $\beta_i$ in the baseline model. Specially, if all $\alpha_i$ and $\beta_i$ are equal to 1, the model will become the baseline model, and the method of feature fusion can be represented by (1). In our three experiments, we set $\alpha_i$=0.1, $\alpha_i$=0.5, and $\beta_i$=0.1 respectively. What's more, the baseline model improved by dynamic weight and gating mechanism is also in the list. Dynamic weight means add a dynamic channel weight for the feature maps comparing to the stable weight in weight mechanism. Dynamic weight is equivalent to the gate mechanism lacking an activation function. Fig. 3 illustrates the concatenation using gating mechanism, and it can be represented by (3), $L$=1.

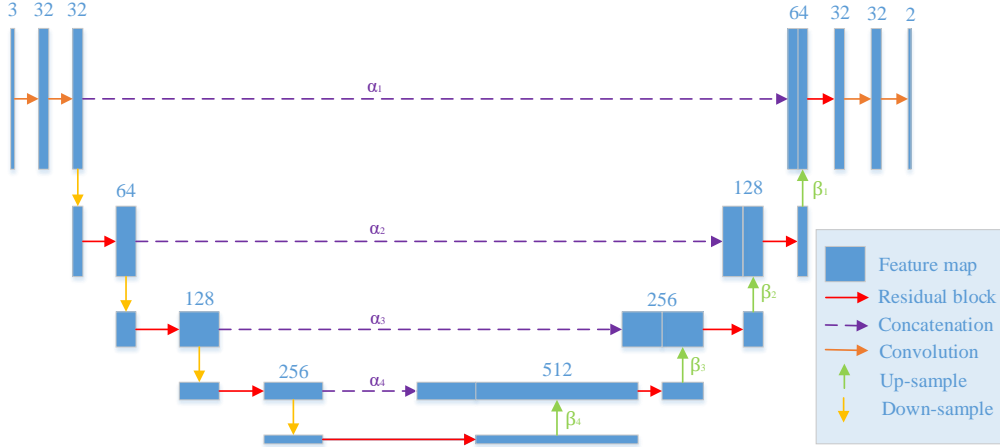

Fig. 2 Illustration of the overall architecture of residual U-Net. $\alpha_i$ and $\beta_i$ are the weights, $i \in \{1, …, 4\}$.

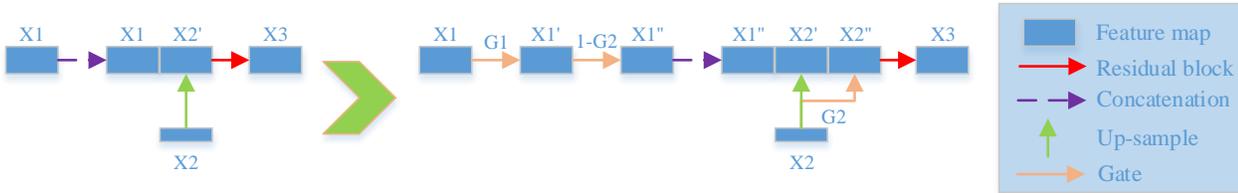

Fig. 3 Illustration of the change from normal concatenation to the concatenation using gating mechanism. $G_i$=$sigmoid(w_i*X_i)$, $i \in \{1, 2\}$.

## IV. EXPERIMENT

In this section, we first introduce the details of the SS model setting. Then, we evaluate weight mechanism in Massachusetts buildings dataset and observe the effect of the weight for U-Net by computing the mean and square error of the parameters in each layer.

### A. Implementation Details

Our implementation is based on Pytorch [20]. All the models are trained on a GTX 1080Ti GPU and a GTX 2080Ti GPU.

**Training settings:** Data augmentation contains random cropping (from 250×250 to 125×125), random scaling in the range of [0.5, 2], and random horizontal flipping. The loss function is cross entropy. SGD-Momentum optimizer with the base learning rate of 0.001, the momentum of 0.9 and the weight decay of 0.0005 is used. We use the poly learning rate policy which learning rate is decayed by $(1-\frac{iter}{max\_iter})^{power}$ with $power$=0.9. 180K training iterations with batch size of 16 is carried for training.

## B. Massachusetts Buildings Dataset

The Massachusetts buildings dataset [4] is applied in the experiments. It contains 151 aerial images of size 1500×1500, including 137 images for training, 4 images for validation and 10 images for testing. The resolution of the images is 1 meter per pixel. The channels of the images are red, green and blue. Considering the computer burden, each image is cut into size of 250×250 pixels.

Table I provides the comparison of our methods with baseline model and the baseline model improved by dynamic weight and gating mechanism on the Massachusetts building test dataset in terms of mean Intersection over Union (mIoU), # of parameters, GFLOPs, pixel accuracy, and mean accuracy. All models are same in training strategies and parameter setting. In the models improved by weight mechanism, the sole changed parameter is $\alpha_i$ or $\beta_i$ illustrated in Fig. 2. In baseline model, $\alpha_i=1$, $\beta_i=1$. In the contrast experiments, the weights are set as $\alpha_i=0.1$ and $\beta_i=1$, $\alpha_i=0.5$ and $\beta_i=1$, $\alpha_i=1$ and $\beta_i=0.1$ respectively.

TABEL I
SEMANTIC SEGMENTATION EXPERIMENTAL RESULTS ON MASSACHUSETTS BUILDINGS DATASET

| Initialization of $W_{BN}$ | Model | # of Parameters | GFLOPs | mIoU | Pixel acc. | Mean acc. |
|---|---|---|---|---|---|---|
| | Baseline | 5.11M | 5.62 | 0.7865 | 0.9265 | 0.8630 |
| | $\alpha_i=0.1$ | +4 | 5.62 | 0.7941 | 0.9296 | 0.8680 |
| | $\alpha_i=0.5$ | +4 | 5.62 | 0.7945 | 0.9297 | 0.8687 |
| $W_{BN} \sim N(0,1)$ | $\beta_i=0.1$ | +4 | 5.62 | 0.7747 | 0.9241 | 0.8436 |
| | Dynamic weight | +480 | 5.63 | 0.7847 | 0.9273 | 0.8539 |
| | Gating mechanism | +0.34M | 6.18 | 0.7881 | 0.9268 | 0.8661 |

Fig. 4 illustrates the distribution of the test results about baseline and changed models to clarify the effect of the weighted concatenation in model training. Each model is trained and tested 10 times independently.

It's clearly shown that adding suitable weights as either $\alpha_i=0.1$ or $\alpha_i=0.5$ in concatenation of series connection gets more precise results comparing to baseline model. However, if adding weight in wrong position like $\beta_i=0.1$, the result will be worse. Dynamic weight for each channel in the feature maps is not helpful for feature fusion. Gating mechanism will improve the outcome but also enlarge the number of the parameters and computation at the same time.

In other words, we just increase four hyperparameters in residual U-Net and limited computation which can be ignored, and achieve 0.8% mIoU improvement comparing to baseline. The shortcoming of our method is obvious, the outcomes of 10 test results are more discrete.

In order to observe the influence of the weight mechanism for the model more clearly, Fig. 5 illustrates the mean value and square error of the model parameters in each layer. Fig. 5(a)-(h) show that $\alpha_i$ and $\beta_i$ can affect the mean value of the model convolution layer weights but basically can't influence the square error. Fig. 5(q)-(x) demonstrate that the bias has limited relationship with $\alpha_i$, but swings seriously with the effect of $\beta_i$. Owing to the normal distribution initialization of weight in BN, namely $W_{BN}$, it's confusing to notice the discipline by inspecting the Fig. 5(i)-(p).

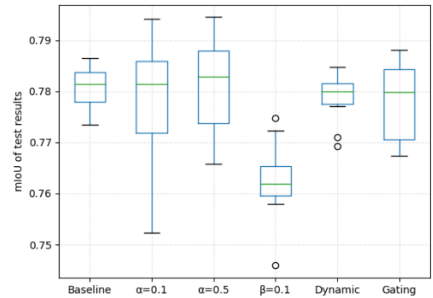

Fig. 4 Box plot of the mIoU of 10 test results about baseline and changed model, and the initialization of $W_{BN}$ is normal distribution.

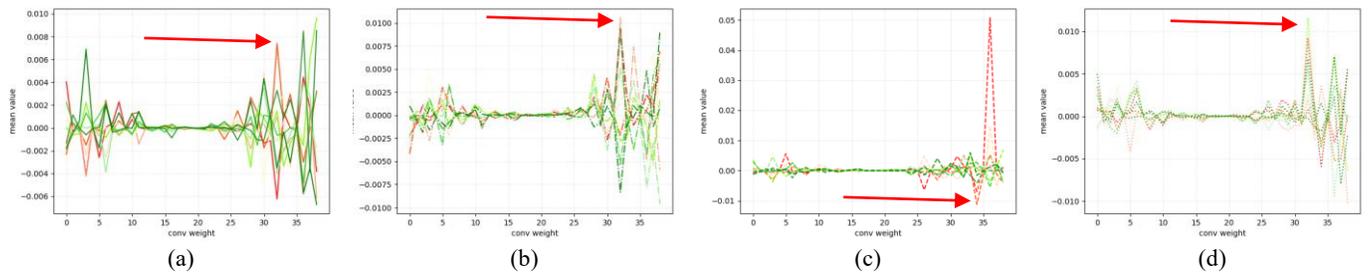

(a)      (b)      (c)      (d)

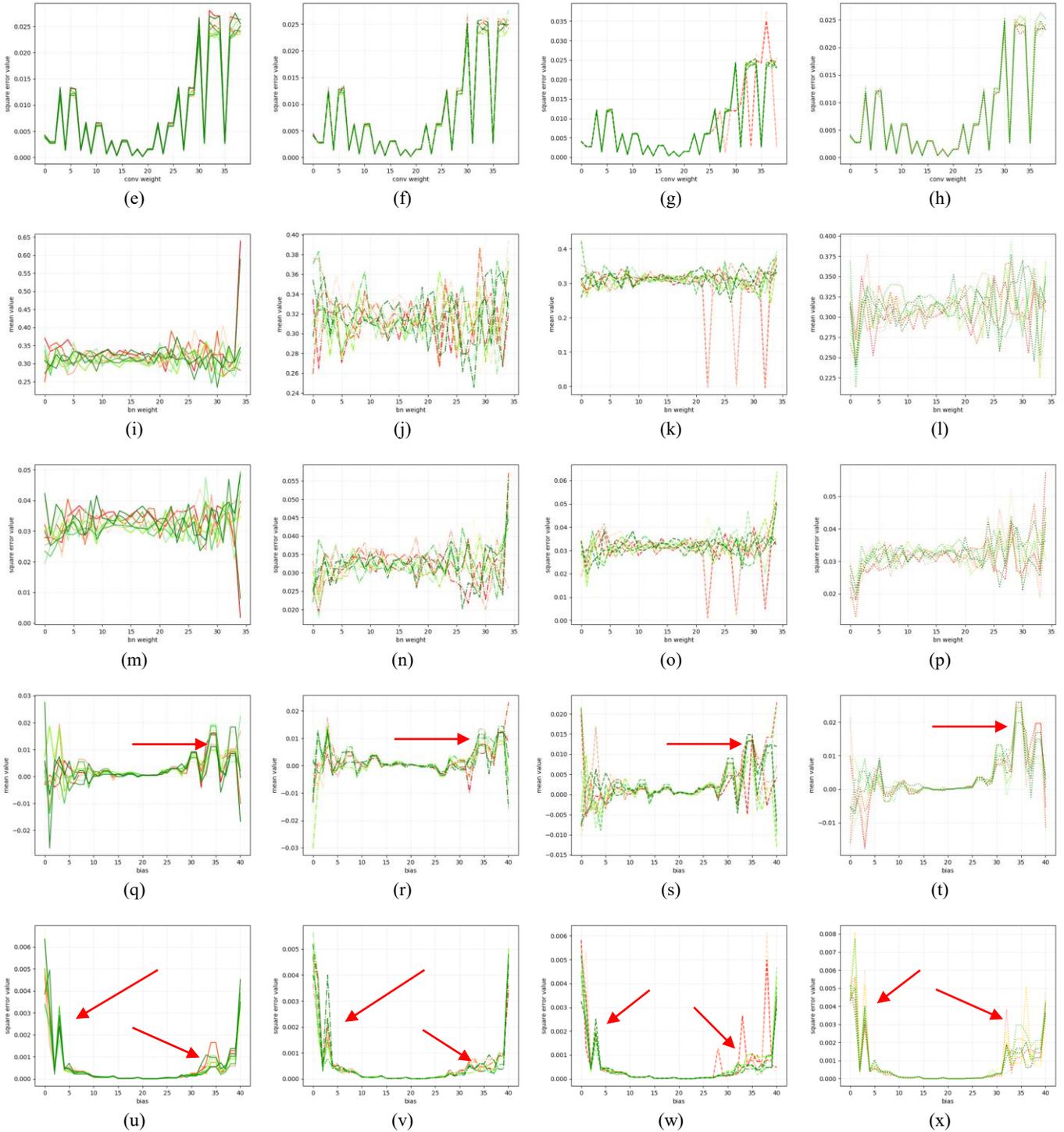

Fig. 5 Illustration of the mean and square error value for model parameters. Solid line, chain line, dashed line and dotted line denote the value of baseline, the model with $α_i$=0.1, $α_i$=0.5, and $β_i$=0.1 respectively. The mIoU of the test result is higher if the color of the line is closer to red. (a), (b), (c), and (d) show the mean value of the convolution layer. (e), (f), (g), and (h) show the square error correspondingly. (i), (j), (k), and (l) show the mean value of $W_{BN}$. (m), (n), (o), and (p) show the square error correspondingly. (q), (r), (s), and (t) show the mean value of bias in the BN layer and the convolution layer which don't belong to the residual block. (u), (v), (w), and (x) show the square error correspondingly.

Thus, we set the initialization of $W_{BN}$ to 1 for better viewing. Table II shows the result of models tested on Massachusetts building dataset, and the initialization of the weights in BN layers is constant 1. Fig. 6 illustrates the distribution of 10 test results about baseline and changed model. It shows that our method can still effectively improve the performance of residual U-Net. The mean value and the square error of $W_{BN}$ in each layer are illustrated in Fig. 7. It shows that $\beta_i$ will affect $W_{BN}$ jitter, whereas under the influence of $\alpha_i$, the change of $W_{BN}$ compared with baseline is not so severe.

TABEL II
SEMANTIC SEGMENTATION EXPERIMENTAL RESULTS ON MASSACHUSETTS BUILDINGS DATASET

| Initialization of $W_{BN}$ | Model | mIoU | Pixel acc. | Mean acc. |
|---|---|---|---|---|
| | Baseline | 0.7820 | 0.9255 | 0.8561 |
| $W_{BN}=1$ | $\alpha_i=0.1$ | 0.7846 | 0.9279 | 0.8506 |
| | $\alpha_i=0.5$ | 0.7900 | 0.9285 | 0.8621 |
| | $\beta_i=0.1$ | 0.7716 | 0.9218 | 0.8468 |

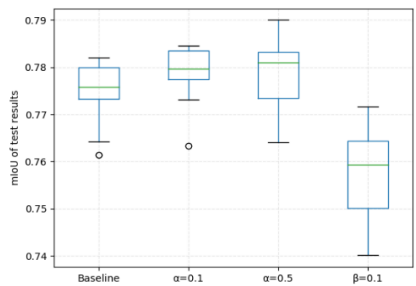

Fig. 6 Box plot of the mIoU of 10 test results about baseline and changed model, and the initialization of $W_{BN}$ is constant 1.

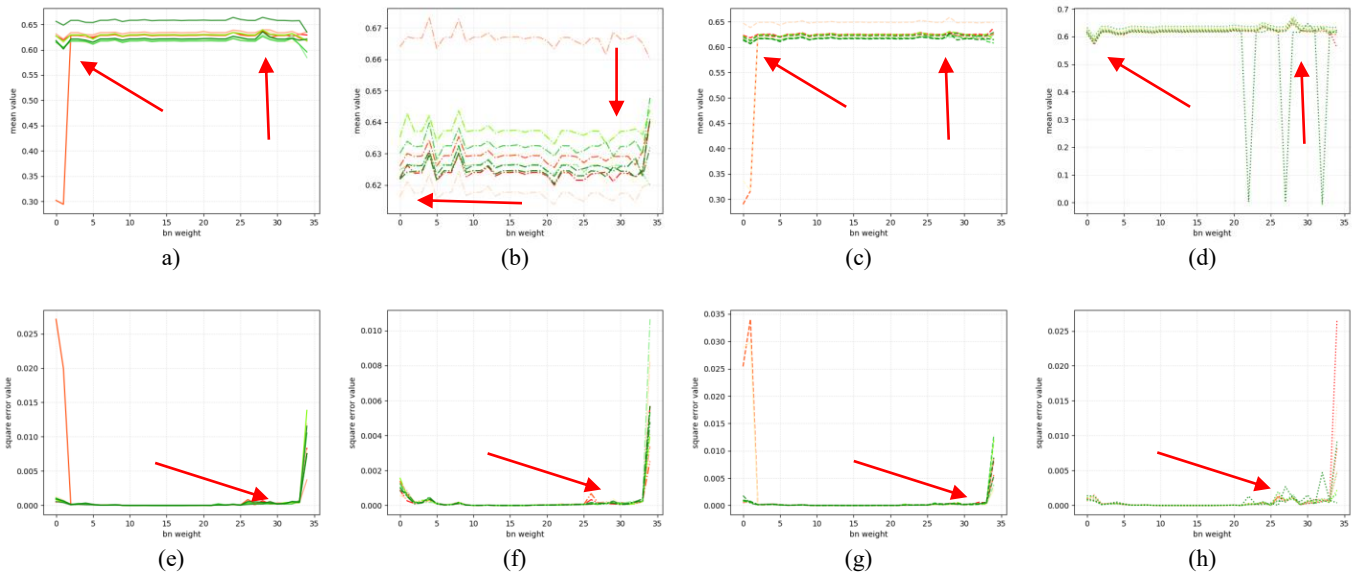

Fig. 7 Illustration of the mean and square error value for $W_{BN}$. Solid line, chain line, dashed line and dotted line denote the value of baseline, the model with $\alpha_i=0.1$, $\alpha_i=0.5$, and $\beta_i=0.1$ respectively. The mIoU of the test result is higher if the color of the line is closer to red. (a), (b), (c), and (d) show the mean value of $W_{BN}$. (e), (f), (g), and (h) show the square error correspondingly.

*C. Fused U-Net*

In order to verify the extensiveness of the application weight mechanism, we design a new network architecture called fused U-Net. The architecture is illustrated in Fig. 8.

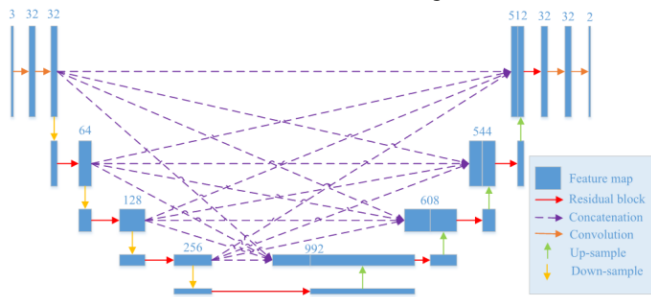

Fig. 8 Illustration of Fused U-Net.

In the comparative experiments, we set the weight to 0.5 and 1 respectively, and train each model 10 times independently. All the training settings are same as section IV.A. The test results of each model is illustrated in Fig. 9. When the weight is equal to 0.5, the best test result is 80.15% mIoU. However, the best test result is 80.03% when the weight is equal to 1. Moreover, when the weight is equal to 0.5, the mIoU of 5 test results is higher than 79.50%, but when the weight is equal to 1, the test results exceed 79.50% mIoU only once.

It clearly shows that proper use of weight mechanism can effectively improve the accuracy of the model.

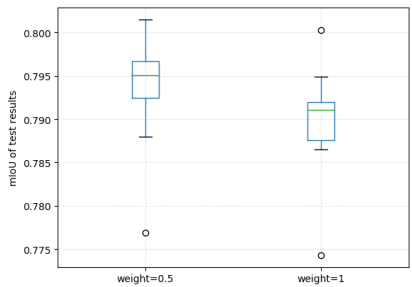

Fig. 9 Box plot of the mIoU of 10 test results about the model with weights of 0.5 and 1.

## V. Conclusion

Different from the naive concatenation and dynamic gating mechanism, we propose weight mechanism for concatenation of series connection, and ensure that suitable weight can improve the effectiveness of concatenation by scaling the combined feature maps. We explore the weight mechanism in residual U-Net and get a better result comparing to baseline model and gating mechanism. The advantage of weight mechanism is that it doesn't change the architecture of the baseline model but just changes the weight in concatenation of series connection to get a better result, which is simple and effective. Meanwhile, we design fused U-Net to test weight mechanism, it shows that weight mechanism can be applied broadly in various models if they have concatenation of series connection.

## Acknowledgment

Thanks Anchang Qi and Yuxiang Xu for sponsoring.

## References

[1] R. K. Srivastava, K. Greff, and J. Schmidhuber, "Training very deep networks," in Neural Information Processing Systems (NIPS), Montreal, 2015.
[2] H. Xu, Y. Gao, F. Yu and T. Darrell, "End-to-end learning of driving models from large-scale video datasets," in 2017 IEEE Conference on Computer Vision and Pattern Recognition (CVPR), Honolulu, 2017, pp. 3530-3538.
[3] X. Li, H. Zhao, L. Han, Y. Tong, and K. Yang, "GFF: gated fully fusion for semantic segmentation," arXiv preprint arXiv: 1904.01803. 2019.
[4] V. Mnih. "Machine learning for aerial image labeling," PhD thesis, University of Toronto, 2013.
[5] K. He, X. Zhang, S. Ren and J. Sun, "Deep residual learning for image recognition," in 2016 IEEE Conference on Computer Vision and Pattern Recognition (CVPR), Las Vegas, 2016, pp. 770-778.
[6] K. He, X. Zhang, S. Ren, and J. Sun, "Identity mappings in deep residual networks," arXiv preprint arXiv:1603.05027. 2016.
[7] G. E. Hinton, N. Srivastava, A. Krizhevsky, I. Sutskever, and R. R. Salakhutdinov, "Improving neural networks by preventing co-adaptation of feature detectors," arXiv preprint arXiv:1207.0580. 2012.
[8] J. Long, E. Shelhamer and T. Darrell, "Fully convolutional networks for semantic segmentation," in 2015 IEEE Conference on Computer Vision and Pattern Recognition (CVPR), Boston, 2015, pp. 3431-3440.
[9] O. Ronneberger, P. Fischer, and T. Brox, "U-net: convolutional networks for biomedical image segmentation," in International Conference on Medical Image Computing and Computer Assisted Intervention (MICCAI), Munich, 2015, pp. 234-241
[10] G. Huang, Z. Liu, L. v. d. Maaten and K. Q. Weinberger, "Densely connected convolutional networks," in 2017 IEEE Conference on Computer Vision and Pattern Recognition (CVPR), Honolulu, 2017, pp. 2261-2269.
[11] Z. Zhou, M. M. R. Siddiquee, N. Tajbakhsh, and J. Liang, "Unet++: A nested u-net architecture for medical image segmentation," in International Conference on Medical Image Computing and Computer Assisted Intervention (MICCAI), Granada, 2018, pp. 3-11.
[12] M. Yang, K. Yu, C. Zhang, Z. Li and K. Yang, "Denseaspp for semantic segmentation in street scenes," in 2018 IEEE Conference on Computer Vision and Pattern Recognition (CVPR), Salt Lake City, 2018, pp. 3684-3692.
[13] L. Chen, G. Papandreou, I. Kokkinos, K. Murphy and A. L. Yuille, "DeepLab: semantic image segmentation with deep convolutional nets, atrous convolution, and fully connected crfs," in IEEE Transactions on Pattern Analysis and Machine Intelligence, vol. 40, no. 4, pp. 834-848, 1 April 2018.
[14] H. Zhao, J. Shi, X. Qi, X. Wang and J. Jia, "Pyramid scene parsing Network," in 2017 IEEE Conference on Computer Vision and Pattern Recognition (CVPR), Honolulu, 2017, pp. 6230-6239.
[15] Y. Zhou, X. Hu, and B. Zhang, "Interlinked convolutional neural networks for face parsing," in International Society of Nutrigenetics and Nutrigenomics (ISNN), Chapel Hill, 2015, pp. 222-231.
[16] K. Sun, Y. Zhao, B. Jiang, T. Cheng, B. Xiao, D. Liu, Y. Mu, X. Wang, W. Liu, and J. Wang, "High-resolution representations for labeling pixels and regions," arXiv preprint arXiv: 1904.04514. 2019.
[17] S. Hochreiter and J. Schmidhuber, "Long Short-Term Memory," in Neural Computation, vol. 9, no. 8, pp. 1735-1780, 15 Nov. 1997.
[18] R. K. Srivastava, K. Greff, J. Schmidhuber, "Highway networks," arXiv preprint arXiv:1505.00387. 2015.
[19] S. Ioffe, and C. Szegedy, "Batch normalization: accelerating deep network training by reducing internal covariate shift," in International Conference on Machine Learning (ICML), Lille, 2015, pp. 448-456.
[20] A. Paszke, S. Gross, S. Chintala, G. Chanan, E. Yang, Z. DeVito, Z. Lin, A. Desmaison, L. Antiga, and A. Lerer, "Automatic differentiation in pytorch," in Neural Information Processing Systems-Workshop (NIPS-W), Long Beach, 2017.